\documentclass[10pt]{article}
\usepackage{fancyhdr}
\usepackage{extramarks}
\usepackage{amsmath}
\usepackage{amsthm}
\usepackage{amsfonts}
\usepackage{siunitx}
\usepackage{tikz}
\usepackage[plain]{algorithm}
\usepackage{algpseudocode}
\usepackage{multirow}
\usepackage{booktabs}
\usepackage{graphicx}
\usepackage{subfigure}
\usepackage[colorlinks,linkcolor=black,anchorcolor=black,citecolor=black,urlcolor=blue]{hyperref}
\usepackage{amsmath,bm}
\usepackage{booktabs}
\usepackage{mathtools}
\usepackage{amssymb}
\usepackage{caption}
\usepackage{capt-of}
\usepackage{mciteplus}
\usepackage{cite}
\usepackage{mathrsfs}
\usepackage[title,titletoc,toc]{appendix}
\usepackage{xr}
\usepackage{parskip}
\usepackage{soul}
\usepackage{textcomp}
\usepackage[colaction]{multicol}
\usepackage[switch]{lineno}
\usepackage{lipsum}
\usepackage{etoolbox}
\usepackage{longtable}
\usepackage{array}
\usepackage{tablefootnote}
\usepackage{ragged2e}
\usepackage{soul}
\newcolumntype{C}[1]{>{\centering\arraybackslash}p{#1}}
\usetikzlibrary{automata,positioning}
\usepackage{enumitem}
\usetikzlibrary{automata,positioning}

\usepackage{xr}

\begin{document}
	\title{Analyzing Single Cell RNA Sequencing with Topological Nonnegative Matrix Factorization}
	
	\author{Yuta Hozumi$^1$ and  Guo-Wei Wei$^{1,2,3}$\footnote{
			Corresponding author.		Email: weig@msu.edu} \\% Author name
		\\
		$^1$ Department of Mathematics, \\
		Michigan State University, East Lansing, MI 48824, USA.\\
		$^2$ Department of Electrical and Computer Engineering,\\
		Michigan State University, East Lansing, MI 48824, USA. \\
		$^3$ Department of Biochemistry and Molecular Biology,\\
		Michigan State University, East Lansing, MI 48824, USA. \\
	}
	\date{\today} % Date for the report

	\maketitle
	
	\begin{abstract}
	Single-cell RNA sequencing (scRNA-seq) is a relatively new technology that has stimulated enormous interest in statistics, data science, and computational biology due to the high dimensionality, complexity, and large scale associated with scRNA-seq data. Nonnegative matrix factorization (NMF) offers a unique approach due to its meta-gene interpretation of resulting low-dimensional components. However, NMF approaches suffer from the lack of multiscale analysis. This work introduces two persistent Laplacian regularized NMF methods, namely, topological NMF (TNMF) and robust topological NMF (rTNMF). By employing a total of 12 datasets, we demonstrate that the proposed TNMF and rTNMF significantly outperform all other NMF-based methods. We have also utilized TNMF and rTNMF for the visualization of popular Uniform Manifold Approximation and Projection (UMAP) and t-distributed stochastic neighbor embedding (t-SNE).

	\end{abstract}
	keywords: Algebraic topology,  Persistent Laplacian, scRNA-seq, dimensionality reduction,   machine learning

\newpage

\section{Introduction}

Single-cell RNA sequencing (scRNA-seq) is a relatively new technology that has unveiled the heterogeneity within cell populations, providing valuable insights into complex biological interactions and pathways, such as cell-cell interactions, differential gene expression, signal transduction pathways, and more \cite{lun2016step}.

Unlike traditional microarray analysis, often referred to as bulk sequencing, scRNA-seq offers the transcriptomic profile of individual cells. With current technology, it's possible to sequence more than 20,000 genes and 10,000 samples simultaneously. Standard experimental procedures involve cell isolation, RNA extraction, sequencing, library preparation, and data analysis.

Over the years, numerous data analysis pipelines have been proposed, typically encompassing data preprocessing, batch correction, normalization, dimensionality reduction, feature selection, cell type identification, and downstream analyses to uncover relevant biological functions and pathways \cite{hwang2018single, andrews2021tutorial, luecken2019current, chen2019single, petegrosso2020machine}.

However, scRNA-seq data, in addition to their high dimensionality, are characterized by nonuniform noise, sparsity due to drop-out events and low reading depth, as well as unlabeled data \cite{lahnemann2020eleven}. Consequently, dimensionality reduction and feature selection are essential for successful downstream analysis.

Principal components analysis (PCA), uniform manifold approximation and projection (UMAP), and t-distributed stochastic neighbor embedding (t-SNE) are among the most commonly used dimensionality reduction tools for scRNA-seq data. PCA is often employed as an initial step in analysis pipelines, such as trajectory analysis and data integration \cite{la2018rna, bergen2020generalizing, luecken2022benchmarking, stuart2019comprehensive}. In PCA, the first few components are referred to as the principal components, where the variance of the projected data is maximized. In PCA, each $i$th component is orthogonal to all the $i-1$ components, maximizing the residual data projected onto the $i$th component \cite{dunteman1989principal, jolliffe2016principal}. Numerous successful extensions to the original formulation have been proposed \cite{cottrell2023plpca, lounici2013sparse, zou2006sparse, townes2019feature}. However, due to the orthogonality constraint of PCA, the reduced data may contain negative values, making it challenging to interpret.

UMAP and t-SNE are nonlinear dimensionality reduction methods often used for visualization. UMAP constructs a $k$-dimensional weighted graph based on $k$-nearest neighbors and computes the edge-wise cross-entropy between the embedded low-dimensional weighted graph representation, utilizing the fuzzy set cross-entropy loss function \cite{mcinnes2018umap}. t-SNE computes the pairwise similarity between cells by constructing a conditional probability distribution over pairs of cells. Then, a student t-distribution is used to obtain the probability distribution in the embedded space, and the Kullback-Leibler (KL) divergence between the two probability distributions is minimized to obtain the reduced data \cite{hinton2002stochastic, van2008visualizing, kobak2021initialization, becht2019dimensionality}. However, due to the stochastic nature of these methods and their instability at dimensions greater than 3 \cite{hozumi2022ccp}, they may not be suitable for downstream analysis.

Nonnegative matrix factorization (NMF) is another dimensionality reduction method in which the objective is to decompose the original count matrix into two nonnegative factor matrices \cite{lee2000algorithms, wang2012nonnegative}. The resulting basis matrices are often referred to as meta-genes and represent nonnegative linear combinations of the original genes. Consequently, NMF results are highly interpretable. However, the original formulation employs a least-squares optimization scheme, making the method susceptible to outlier errors \cite{liu2006nonnegative}.

To address this issue, Kong et al. \cite{kong2011robust} introduced robust NMF (rNMF), or $l_{2,1}$-NMF, which utilizes the $l_{2,1}$-norm and can better handle outliers while maintaining comparable computational efficiency to standard NMF. Manifold regularization has also been employed to incorporate geometric structures into dimensionality reduction, utilizing a graph Laplacian, leading to Graph Regularized NMF (GNMF) \cite{xiao2018graph}. Semi-supervised methods, such as those incorporating marker genes \cite{wu2020robust}, similarity and dissimilarity constraints \cite{shu2022robust}, have been proposed to enhance NMF's robustness. Additionally, various other NMF derivatives have been introduced \cite{lan2022detecting, liu2017joint, yu2019robust}.

Despite these advancements in NMF, manifold regularization remains an essential component to ensure that the lower-dimensional representation of the data can form meaningful clusters. However, using graph Laplacians can only capture a single scale of the data, specifically the scaling factor in the heat kernel. Therefore, single-scale graph Laplacians lack multiscale information.

 Eckmann et al. \cite{eckmann1944harmonische} introduced simplicial complexes to the graph Laplacian defined on point cloud data, leading to the combinatorial Laplacian. This can be viewed as a discrete counterpart of the de Rham-Hodge Laplacian on manifolds. Both the Hodge Laplacian and the combinatorial Laplacian are topological Laplacians that give rise to topological invariants in their kernel space, specifically the harmonic spectra. However, the nonharmonic spectra contain algebraic connectivity that cannot be revealed by the topological invariants \cite{horak2013spectra}.

A significant development in topological Laplacians occurred in 2019 with the introduction of persistent topological Laplacians. Specifically, evolutionary de Rham theory was introduced to obtain persistent Hodge Laplacians on manifolds \cite{chen2021evolutionary}. Meanwhile, persistent combinatorial Laplacian \cite{wang2020persistent}, also known as the persistent spectral graph or persistent Laplacian (PL), was introduced for point cloud data. These methods have spurred numerous theoretical developments \cite{memoli2022persistent, liu2023algebraic, wei2021persistent, wang2023persistent, chen2023persistent} and code construction \cite{wang2021hermes}, as well as remarkable applications in various fields, including protein engineering \cite{qiu2023persistent}, forecasting emerging SARS-CoV-2 variants BA.4/BA.5 \cite{chen2022persistent}, and predicting protein-ligand binding affinity \cite{meng2021persistent}. Recently, PL has been shown to improve PCA performance \cite{cottrell2023plpca, cottrell2023knnplpca}.

This growing interest arises from the fact that persistent topological Laplacians represent a new generation of topological data analysis (TDA) methods that address certain limitations of the popular persistent homology \cite{zomorodian2004computing, edelsbrunner2008persistent}. In persistent homology, the goal is to represent data as a topological space, often as simplicial complexes. Then, ideas from algebraic topology, such as connected components, holes, and voids, are used to extract topological invariants during a multiscale filtration. Persistent homology has facilitated topological deep learning (TDL), an emerging field \cite{cang2017topologynet}. However, persistent homology is unable to capture the homotopic shape evolution of data. PLs overcome this limitation by tracking changes in non-harmonic spectra, revealing the homotopic shape evolution. Additionally, the persistence of PL's harmonic spectra recovers all topological invariants from persistent homology.

In this work, we introduce PL-regularized NMF, namely the topological NMF (TNMF) and robust topological NMF (rTNMF). Both TNMF and rTNMF can better capture multiscale geometric information than the standard GNMF and rGNMF. To achieve improved performance, PL is constructed by observing cell-cell interactions at multiple scales through filtration, creating a sequence of simplicial complexes. We can then view the spectra at each complex associated with a filtration to capture both topological and geometric information. Additionally, we introduce $k$-NN based PL to TNMF and rTNMF, referred to as $k$-TNMF and $k$-rTNMF, respectively. The $k$-NN based PL reduces the number of hyperparameters compared to the standard PL algorithm.

The outline of this work is as follows. First, we provide a brief overview of NMF, rNMF, GNMF, and rGNMF. Next, we present a concise theoretical formulation of PL and derive the multiplicative updating scheme for TNMF and rTNMF. Additionally, we introduce an alternative construction of PL, termed $k$-NN PL. Following that, we present a benchmark using 12 publicly available datasets. We have observed that PL can improve NMF performance by up to 0.16 in ARI, 0.08 in NMI, 0.04 in purity, and 0.1 in accuracy.

\section{Methods}
In this section, we provide a brief overview of NMF methods, namely NMF, rNMF, GNMF, and rGNMF. We then give persistent Laplacian and its construction. Finally, we formulate various PL   regularized NMF methods.

\subsection{Prior Work}
\paragraph{NMF}
The original formulation of NMF utilizes the Frobenius norm, which assumes that the noise of the data is sample from Gaussian distribution.
\begin{align}
    \min_{W, H}\|X - WH\|_F^2, \quad \text{s.t. } W, H \ge 0
\end{align}
where $\|A\|_F^2 = \sum_{i,j}a_{ij}^2$.
Lee et al. proposed a multiplicative updating scheme, which preserves the nonnegativity \cite{lee2000algorithms}. For the $t+1$th iteration,
\begin{align}
    & w^{t+1} = w_{ij}^t \frac{(XH^T)_{ij}}{(WHH^T)_{ij}} \\
    & h^{t+1} = h_{ij}^t \frac{(W^TX)_{ij}}{(W^TWH)_{ij}}
\end{align}
Although the updating scheme is simple and effective in many biological data applications, scRNA-seq data is sparse and contains large amount of noise. Therefore, a model that is more robust to noise is necessary for feature selection and dimensionality reduction

\paragraph{rNMF}
The robust NMF (rNMF) utilizes the $l_{2,1}$ norm, which assumes that the noise of the data is sampled from a Laplace distribution, which may be more suitable for a count-based data matrix, like scRNA-seq. The minimization function is given as the following
\begin{align*}
    \min_{W,H}\| X - WH\|_{2,1},  \quad \text{s.t. } W, H \ge 0,
\end{align*}
where $\|A\|_{2,1} = \sum_j \|\mathbf{a}_j\|_2$. Because $l_{2,1}$-norm utilizes summation over the $l_2$ distance of the original cell feature and the reduced feature, the effect of the outlier will not dominate the loss function as much as the Frobenius norm formulation. RNMF has the following updating scheme
\begin{align}
    & w^{t+1}_{ij} = w_{ij}^{t} \frac{(XQH^T)_{ij}}{(WHQH^T)_{ij}} \\
    & h_{ij}^{t+1} = h_{ij}^t \frac{(W^TXQ)_{ij}}{(W^TWHQ)_{ij}},
\end{align}
where $Q_{jj} = 1/\|X - W\mathbf{h}_j\|_2$.

\paragraph{GNMF amd rGNM}
Manifold regularization has been widely utilized in scRNA-seq. Let $G(V, E, W)$ be a graph, where $V = \{\mathbf{x}_j\}_{j=1}^N$ is the set of vertices, $E = \{(\mathbf{x}_i, \mathbf{x}_j) | \mathbf{x}_i \in \mathcal{N}_k (\mathbf{x}_j) \cup \mathbf{x}_j \in \mathcal{N}_k (\mathbf{x}_i) \}$ is the set of edges, and $W$ is the weight associated with the edges. Here, $\mathcal{N}_k(\mathbf{x}_j)$ denotes the $k$-th nearest neighbors of vertex $j$.   The heat kernel is often used to construct the weight, and we can construct the adjacency matrix $A$ as the following.
\begin{align}
    A_{ij} = \begin{cases}
        \exp\left(-\frac{\|\mathbf{x}_i - \mathbf{x}_j\|^2}{\sigma}\right) & \mathbf{x}_j \in \mathcal{N}_k (\mathbf{x}_i)\\
        0, & \text{otherwise}.
    \end{cases}
\end{align}
Since heat kernel satisfies the conditions $W_{ij} \to 0$ as $\|\mathbf{x}_i - \mathbf{x}_j\| \to \infty$ and $W_{ij} \to 1$ as $\|\mathbf{x}_i - \mathbf{x}_j\| \to 0$, we can construct the graph regularization term, $R_G$, by looking at the distance $\|\mathbf{h}_i - \mathbf{h}_j\|^2$.
\begin{align*}
    R_G & = \frac{1}{2}\sum_{i,j} A_{ij} \|\mathbf{h}_i - \mathbf{h}_j\|^2  \\
    & = \sum_{i}D_{ii}\mathbf{h}_i^T\mathbf{h}_i -\sum_{ij} A_{ij} \mathbf{h}_i^T\mathbf{h}_j \\
    & = \text{Tr}(HDH^T) - \text{Tr}(HAH^T) \\
    & = \text{Tr}(HLH^T).
\end{align*}
Here, $L$ and $D$ are the Laplacian and the degree matrix, given by $L= D - A$ and $D_{ii} = \sum_j A_{ij}$, respectively. $\text{Tr}(\cdot)$ denotes the trace of the matrix. Utilizing the regularization parameters, $\lambda \ge 0$, we get the objective function of GNMF
\begin{align}
    \min_{W,H} \|X - WH\|_F^2 + \lambda \text{Tr}(HLH^T).
\end{align}
and the objective function for rGNMF
\begin{align}
    \min_{W,H} \|X - WH\|_{2,1} + \lambda \text{Tr}(HLH^T).
\end{align}

\subsection{Topological NMF}
While graph regularization improves the traditional NMF and rNMF, the choice of $\sigma$ and vastly change the result. Furthermore, graph regularization only captures a single scale, and may  not be able to capture the mutliscale geometric information in data. Here, we give a brief introduction to persistent homology and persistent Laplacian and derive the updating scheme for the topological NMF.

\subsubsection{Persistent Laplacians}
Persistent homology and persistent spectral graphs have been successfully used in biomolecular data \cite{qiu2023persistent, meng2021persistent, cang2017topologynet, zomorodian2004computing, edelsbrunner2008persistent, cottrell2023plpca}. Similar to persistent homology, persistent spectral graphs track birth and death of topological features, i.e., holes, over different scales. However,  unlike persistent homology, persistent spectral graphs can further capture the homotopic shape evolution of data during the  filtration.  Through the filtration process, these methods offer  the multiscale analysis of  data.

We   begin by the definition of simplex. Let $\sigma_q= [v_0, \cdots,v_q]$ denote $q$-simplex, where $v_i$ is a vertex. $\sigma_0$ is a node, $\sigma_1$ is an edge, $\sigma_2$ is a triangle $\sigma_3$ is a tetrahedron, and so on. A simplicial complex $K$ is a union  of simplicies such that 
\begin{enumerate}
    \item If $\sigma_q \in K$ and $\sigma_p$ is a face of $\sigma_q$, then $\sigma_p \in K$
    \item The nonempty intersection of any 2 simplicies in $K$ is a face of both simplicies.
\end{enumerate}
We can think of $K$ as gluing lower dimensional simplicies that satisfies the above 2 properties.

A $q$-chain is a formal sum of $q$-simplicies in $K$ with the coefficients $\mathbb{Z}_2 = \{0,1\}$. The set of all $q$-chains has contains the basis for the set of $q$-simplicies in $K$. Such set forms a finitely generated free Abelian group $C_q(K)$. We can relate the chain groups via a boundary operator, which is a group homomorphism $\partial_q:C_q(K) \to C_{q-1}(K)$. The boundary operator is defined as the following.
\begin{align}
    \partial_q\sigma_q := \sum_{i=0}^q(-1)^i\sigma_{q-1}^i
\end{align}
where $\sigma_{q-1}^i = [v_0 ,,,,v_i^*, ..., v_q]$, where $\sigma_{q-1}^i$ is a $(q-1)$-simplex with vertex $v_i$ removed. The sequence of chain group connected by the boundary operator defines the chain complex.
\begin{align}
    ...\xrightarrow{\partial_{q+2}} C_{q+1} \xrightarrow{\partial_{q+1}} C_q(K) \xrightarrow{\partial_{q}} ... 
\end{align}
The chain complex associated with a simplicial complex $K$ defines the $q$-th homology group $H_q = \text{Ker}\partial_q/\text{Im}\partial_q$, and the dimension of $H_q$ is the $q$-dimensional holes, or the $q$th Betti number denoted as $\beta_q$. For example, $\beta_0$ is the number of connected components, $\beta_1$ is the number of loops and $\beta_2$ is the number of cavities.

We can now define the dual chain complex through the adjoint operator of $\partial_q$. The dual space is defined as $C^q(K) \cong C_q^*(K)$, and the coboundary operator $\partial_q^*$ is defined as $\partial_q^*:C^{q-1}(K) \to C^q(K)$. For  $\omega^{q-1}\in C^{q-1}(K)$ and $c_q \in C_q(K)$, the coboundary operator is defined as
\begin{align}
    \partial^*\omega^{q-1}(c_q) \equiv \omega^{q-1}(\partial c_q).
\end{align}
Here $\omega^{q-1}$ is a $(q-1)$ cochain, or a homomorphic mapping from a chain to the coefficient group. The homology of the dual chain complex is called the cohomology.

We then define the $q$-combinatorial Laplacian operator $\triangle_q: C^q(K) \to C^q(K)$
\begin{align}
    \triangle_q:= \partial_{q+1}\partial_{q+1}^* + \partial_q^*\partial_q.
\end{align}
Let $\mathcal{B}_q$ be the standard basis for the matrix representation of $q$-boundary operator from $C_q(K)$ and $C_{q-1}(K)$, and $\mathcal{B}_q^T$ be th $q$-coboundary operator. The matrix representation of the $q$-th order Laplacian operator $\mathcal{L}_q$ is defined as 
\begin{align}
    \mathcal{L}_q = \mathcal{B}_{q+1}\mathcal{B}_{q+1}^T + \mathcal{B}_q^T\mathcal{B}_q.
\end{align}
The multiplicity of zero eigenvalue of $\mathcal{L}_q$ is the $q$-th Betti number of the simplicail complex. The nonzero eigenvalues (non-harmonic spectrum) contains other topological and geometrical features.

As stated before, simplicial complex does not provide sufficient information to understand the geometry of the data. To this end, we utilize simplicial complex induced by filtration
\begin{align}
    \{\emptyset\}  = K_0 \subseteq K_1 \subseteq \cdots \subseteq K_p = K,
\end{align}
where $p$ is the number of filtration.

For each $K_t$ $0 \le t \le p$, denote $C_q(K_t)$ as chain group induced by $K_t$, and the corresponding boundary operator $\partial_q^t: C_q(K_t) \to C_{q-1}(K_t)$, resulting in
\begin{align}
    \partial_q^t \sigma_q = \sum_{i=1}^q (-1)^i\sigma_{q-1}^{i-1},
\end{align}
for $\sigma_q \in K_t$. The adjoint operator of $\partial_q^{t}$ is similarity defined as $\partial_q^{t*}: C^{q-1}(K_t) \to C^q(K_t)$, which we regard as the mapping $C_{q-1}(K_t) \to C_{q}(K_t)$ via the isomorphism between cochain and chain groups. Through these 2 operators, we can define the chain complexes induced by $K_t$.

Utilizing filtration with simplicial complex, we can define persistence Laplacian spectra. Let $C_q^{t+p}$ whose boundary is in $C_{q-1}^t$ be $\mathbf{C}_q^{t+p}$, assuming an inclusion mapping $C_{q-1}^t \to C_{q-1}^{t+p}$. On this set, we can define the $p$-persistent $q$-boundary operator denoted $\hat{\partial}_q^{t,p}:\mathbb{C}_q^{t,p} \to C_{q-1}^t$ and the corresponding adjoint operator $(\hat{\partial}^{t,p})^*:C_{q-1}^t \to \mathbb{C}_q^{t,p}$. Then, the $q$-order $p$-persistent Laplacian operator is computed as
\begin{align}
    \triangle_q^{t,p} = \hat{\partial}_{q+1}^{t,p}(\hat{\partial}_{q+1}^{t,p})^* + (\hat{\partial}_q^{t})^*\hat{\partial}_q^{t},
\end{align}
and its matrix representation as
\begin{align}
    \mathcal{L}_q^{t,p} = \mathcal{B}_{q+1}^{t,p}(\mathcal{B}_{q+1}^{t,p})^T + (\mathcal{B}_q^t)^T\mathcal{B}_q^t.
\end{align}
Likewise as before, the multiplicity of the zero-eigenvalue  is the $q$-th order $p$-persistent Betti number $\beta_q^{t,p}$, which is the $q$-dimensional hole in $K_t$ that persists in $K_{t+p}$. Moreover, the $q$-th order Laplacian is just a particular case of $\mathcal{L}_q^{t,p}$, where $p =0$, which is a snapshot of the topology at the filtration step $t$ \cite{wang2021hermes, wang2020persistent}.

We can utilize the 0-persistent Laplacian to capture the interactions between the data at different filtration values. In particular, we can perform filtration by computing a family of subgraphs induced by a threshold distance $r$, which is called the Vietoris Rips complex. Alternatively, we can compute a Gaussian Kernel induced distance to construct the subgraphs.

\subsubsection{TNMF and rTNMF}

For scRNA-seq data, we calculate the 0-persistent Laplacian using the Vietoris-Rips (VR) complexes by increasing the filtration distance. We can then take a weighted sum over the 0-persistent Laplacian induced by the changes in the filtration distance. For persistent Laplacian enhanced NMF, we will provide a computationally efficient algorithm to construct the persistent Laplacian matrix.

Let $L$ be a Laplacian matrix induced by some weighted graph, and note the following
\begin{align*}
    L = \begin{cases}
        l_{ij}, & i \ne j \\
        -\sum_{j=1}^N l_{ij} & i = j.
    \end{cases}
\end{align*}
Then, let $l_{\max} = \max_{i\ne j} l_{ij}$,  $l_{\min} = \min_{i\ne j} l_{ij}$ and $d = l_{\max} - l_{\min}$. The $t$-th Persistent Laplacian $L^t$, $t = 1, ..., T$ is defined as $L^t = \{l_{ij}^t\}$, where 
\begin{align}
    & l_{ij}^t = \begin{cases}
        0 \quad l_{ij} \le (t/T)d + l_{\min} \\
        1 & \text{otherwise}
    \end{cases} \\
    & l_{ii}^t = -\sum_{i\ne j}l_{ij}^t.
\end{align}
Then, we can take the weighted sum over the all the persistent Laplacians 
\begin{align}
    PL := \sum_{t=1}^T \zeta_t L^t.
\end{align}
Unlike the standard Laplacian matrix $L$, PL captures the topological features that persists over different filtration, thus providing a multiscale view of the data that standard Laplacian lacks. Here,  $\zeta_t$ is the hyper-parameter  and must be chosen as a hyperparameter. Then, the topological NMF (TNMF) is defined as 
\begin{align}
    \|X - WH\|_F^2 + \text{Tr}(H^T(PL)H) 
\end{align}
and the topological rNMF (rTNMF) is defined as 
\begin{align}
    \|X-WH\|_{2,1} + \text{Tr}(H^T(PL)H).
\end{align}

\subsubsection{Multiplicative Updating scheme}
The updating scheme follows the same principle as the standard GNMF and rGNMF.

\paragraph{TNMF}
For top-NMF, the Lagrangian function is defined as
\begin{align}
    \mathcal{L} & = \|X - WH\|_F^2 + \lambda  \text{Tr}(H^T(PL)H) + \text{Tr}(\Phi W) + \text{Tr}(\Psi H) \\
                 & = \text{Tr}(X^TX) - 2\text{Tr}(XH^TW^T) + \text{Tr}(WHH^TW^T)+ \lambda  \text{Tr}(H^T(PL)H) + \text{Tr}(\Phi W) + \text{Tr}(\Psi H).
\end{align}
Taking the partial with respect to $W$, we get
\begin{align}
    \frac{\partial \mathcal{L}}{\partial W} = -2H^TXH + 2WHH^T + \Phi.
\end{align}
Using the KKT condition $\Phi_{ij}w_{ij} = 0$, we get the following
\begin{align}
    (-2XH^T)_{ij}w_{ij} +(2WHH^T)_{ij} w_{ij} = 0.
\end{align}
Therefore, the updating scheme is
\begin{align}
    w_{ij}^{t+1} \leftarrow w_{ij}^t \frac{(XH^T)_{ij}}{(WHH^T)_{ij}}.
\end{align}
For updating $H$, we take the derivative of the Lagrangian function with respect to $H$
\begin{align}
    \frac{\partial \mathcal{L}}{\partial H}  = -2 W^TX + 2W^TWH  + 2\lambda H (PL)+ \Psi.
\end{align}
Using the Karush–Kuhn–Tucker (KKT) condition, we have $\Psi_{ij}h_{ij} =0$ and obtain
\begin{align}
     -2(W^TX + \lambda H(PA))_{ij}h_{ij}  + 2(W^TWH + \lambda H(PD))_{ij}h_{ij} = 0,
\end{align}
where $PL = PD - PA$ and $PD_{ii} = \sum_{i\ne j} PA_{ij} $. The updating scheme is then given by
\begin{align}
    h_{ij}^{t+1} \leftarrow h_{ij}^t\frac{(W^TWH + \lambda H(PD))_{ij}}{(W^TX + \lambda H(PA))_{ij}}.
\end{align}

\paragraph{rTNMF}
For the updating scheme for top-rNMF, we utilize the fact that $\|A\|_{2,1} = \text{Tr}(A QA^T)$, where $Q_{ii} = \frac{1}{2\|A_i\|_2}$. The Lagrangian is given by
\begin{align}
    \mathcal{L} & =  \|X - WH\|_{2,1} + \lambda  \text{Tr}(H^T(PL)H) + \text{Tr}(\Phi W) + \text{Tr}(\Psi H) \\
                & = \text{Tr}( (X-WH)Q(X-WH)^T ) + \lambda  \text{Tr}(H^T(PL)H) + \text{Tr}(\Phi W) + \text{Tr}(\Psi H) \\
                & = \text{Tr}( XQX^T ) -2\text{Tr}(WHQ) + \lambda \text{Tr}(H^T(PL)H) + \text{Tr}(\Phi W) + \text{Tr}(\Psi H),
\end{align}
where $Q_{ii} = \frac{1}{\|\mathbf{x}_j - W\mathbf{h}_j\|}$.
Taking the partial with respect to $W$, we get
\begin{align}
    \frac{\partial L}{\partial W} = -(XQH^T) + WHQH^T - \Phi.
\end{align}
Using the KKT conditions $\Phi_{ij}w_{ij} = 0$, we get
\begin{align}
    -(XQH^T)_{ij}w_{ij} + (WHQH^T)_{ij}w_{ij} = 0,
\end{align}
which gives the updating scheme
\begin{align}
    w_{ij}^{t+1} \leftarrow w_{ij}^t \frac{(XQH^T)_{ij}}{ (WHQH^T)_{ij}}.
\end{align}
For $H$, we take the partial with respect to $H$.
\begin{align}
    \frac{\partial L}{\partial H} = -W^TXQ + W^TWHQ + 2\lambda H(PL) + \Psi.
\end{align}
Then, using the KKT conditions $\Psi_{ij}h_{ij} = 0$, we get
\begin{align}
    (-W^TXQ - 2\lambda H(PA))_{ij}h_{ij} + (W^TWHQ + 2\lambda H(PD))_{ij}h_{ij} = 0,
\end{align}
where $PL = PD - PA$ and gives the updating scheme
\begin{align}
    h_{ij}^{t+1}  \leftarrow h_{ij}^t \frac{(W^TXQ+2\lambda H(PA))_{ij}}{(W^TWHQ + 2\lambda H(PD))_{ij}}. 
\end{align}

\subsection{$k$-NN induced Persistent Laplacian}
One major issue with top-GNMF and top-rGNMF is that the parameters $\{\zeta_t\}_{t=1}^T$ have to be chosen. For the parameters, we let $\zeta_t \in \{0, 1, 1/2, \cdots , 1/T\}$ for a total of $T+1$ parameters. Therefore, the number of parameters that needs to be chosen increases exponentially as the number of filtration $T$ increases. Therefore, we propose an approximation to the original formulation using $k$-NN based persistent Laplacian.

Let $\mathcal{N}_t(\mathbf{x}_j)$ be the $t$-nearest neighbors of sample $\mathbf{x}_j$. Then, define the $t$-persistent directed adjacency matrix $\tilde{A}^t$ as
\begin{align}
    \tilde{A}^t = \{\tilde{a}_{ij}^t\}, \quad \tilde{a}_{ij}^t = \begin{cases}
        1 & \mathbf{x_j} \in \mathcal{N}_t(\mathbf{x}_i) \\
        0 &  \text{otherwise}.
    \end{cases}
\end{align}
Then, the $k$-NN based directed adjacency Laplacian is the weighted sum of $\{A^t\}$
\begin{align}
    \tilde{A} := \sum_{t=1}^T \zeta_t \tilde{A}^t.
\end{align}
Then, the undirected persistent adjacency matrix can be obtained via symmetrization
\begin{align*}
    PA = \tilde{A} + \tilde{A}^T - \tilde{A}\cdot \tilde{A}^T,
\end{align*}
where $\cdot$ denote Hadamard product. Then, the persistent Laplacian can be constructed using the persistent degree matrix
\begin{align}
    PL = PD - PA, \quad PD_{ii} = \sum_{j\ne i}PA_{ij}.
\end{align}

One advantage of utilizing the $k$-NN induced persistent Laplacian is that the parameter space is much smaller. We can set $\zeta_t \in \{0, 1\}$, where $\zeta_t = 0$ would 'turn-off' the particular neighbor's connectivity. In essence, the number of parameters will be reduced to $2^T$, a significant decrease from $T(T+1)$ of the original formulation.

\subsection{Evaluation metrics}
Let $Y = \{Y_1, ..., Y_L\}$ and $C = \{C_1, ..., C_L\}$ be 2 partitions of the data. Here, we let $Y$ be the true label partition and $C$ be the cluster label partition. Let $\{y^i\}_{i=1}^N$ and $\{c^i\}_{i=1}^N$ be the true and predicted labels of sample $i$.

\paragraph{Adjusted Rand Index}
Adjusted random index (ARI) measures the similarity between two clustering by observing all pairs of samples that belong to the same cluster, and seeing if the other clustering result also have the same pair of samples in the same cluster \cite{hubert1985comparing}.
Let $n_{ij} = |T_i \cap S_j|$ be the number of samples that belong to true label $i$ and cluster label $j$, and define $a_i = \sum_j n_{ij}$ and $b_{j} = \sum_i n_{ij}$. Then, the ARI is defined as
\begin{align}
    \text{ARI} = \frac{\sum_{ij} \begin{pmatrix}
            n_{ij} \\ 2
        \end{pmatrix} - \left[ \sum_i \begin{pmatrix}
            a_i \\ 2\end{pmatrix} \sum_j \begin{pmatrix}
            b_j \\ 2\end{pmatrix}\right] / \begin{pmatrix}
            N \\ 2\end{pmatrix} }{\frac{1}{2} \left[\begin{pmatrix}
            a_i \\ 2\end{pmatrix} + \begin{pmatrix}
            b_j \\ 2\end{pmatrix}\right] - \left[\begin{pmatrix}
            a_i \\ 2\end{pmatrix} \sum_j \begin{pmatrix}
            b_j \\ 2\end{pmatrix}\right] / \begin{pmatrix}
            N \\ 2\end{pmatrix}}.
\end{align}
The ARI takes on a value between -1 and 1, where 1 is a perfect match between two clustering methods, and 0 is a completely random assignment of labels, and -1 indicates that the two clusterings are completely different.

\paragraph{Normalized Mutual Information}

The normalized mutual information (NMI) measures the mutual information between two clustering results and normalized according to cluster size \cite{vinh2009information}. We fix the true labels $Y$ as one of the clustering result, and use the predicted labels as the other to calculate NMI. The NMI is calculated as the following
\begin{align}
    \text{NMI} = \frac{2I(Y; C)}{H(Y)H(C)},
\end{align}
where $H(\cdot)$ is the entropy and $I(Y;C)$ is the mutual information between true labels $Y$ and predicted labels $C$. NMI has a range of 0 and 1, where 1 is a perfect mutual correlation between the two  sets of labels and 0 means no mutual information.

\paragraph{Accuracy}
Accuracy (ACC) calculates the percentage of correctly predicted class labels. The accuracy is given by
\begin{align}
    \text{ACC} = \frac{1}{N} \sum_{i=1}^N \delta(y^i, f(c^i)),
\end{align}
where $\delta(a,b)$ is the indicator function, where if $a = b$, $\delta(a,b) = 1$, and 0 otherwise. $f:C\to Y$ maps the cluster labels to the true labels, where the mapping is the optimal permutation of the cluster labels and true labels obtained from the Hungarian algorithm \cite{crouse2016implementing}.

\paragraph{Purity}
For purity calculation, each predicted label $C_i$ is assigned to a true label $Y_j$ such that the $|C_i \cap Y_j|$ is maximized \cite{rao2018exploring}. Taking the average over all the predicted label, we obtain the following
\begin{align}
    \text{Purity} = \frac{1}{N}\max_j |C_i \cap Y_j|.
\end{align}
Note that unlike accuracy, purity does not map the predicted labels to the true labels.

\section{Results}

\subsection{Benchmark Data}
We have performed benchmark on 12 publicly available datasets. The GEO accession number, reference, organism, number of cell types, and number of samples are recorded in \autoref{tab: data}. For each data, cell types with less than 15 cells were removed. Log-normalization was applied, and scaled the data to have unit length. For GNMF and rGNMF, $ k =8$ neighbors were used. For TNMF and rTNMF, 8 filtration values were used to construct PL, and for each scale, binary selection $\zeta_p = \{0,1\}$ was used. for $k$-TNMF and $k$-rTNMF, $k=8$ was used with $\zeta_p = \{0,1\}$. For each test, double nonnegative singular value decomposition with zeros filled with the average of $X$ (NNDSVDA) was used for the initialization. The $k$-means clustering was applied to obtain the clustering results.

\begin{table}[H]
    \centering
    \caption{GEO accession code, reference, organism type, cell type, number of samples, and number of genes of each dataset.}
    \begin{tabular}{|c|c c c c c|} \hline
        Geo Accession & Reference & Organism & Cell type & Number of Samples & Number of Genes\\\hline
        GSE67835 & Dramanis \cite{darmanis2015survey} & Human & 8& 420 & 22084\\
        GSE75748 time & Chu \cite{chu2016single} & Human & 6 & 758 & 19189 \\
        GSE82187 & Gokce \cite{gokce2016cellular} & Mouse & 8 & 705 & 18840\\
        GSE84133human1 & Baron \cite{baron2016single} & Human &  9 & 1895  & 20125\\
        GSE84133human2 & Baron \cite{baron2016single}& Human&  9 & 1702 & 20125\\
        GSE84133human3 & Baron \cite{baron2016single} & Human & 9 & 3579 & 20125\\
        GSE84133human4 & Baron \cite{baron2016single} & Human  &  6 & 1275 & 20125\\
        GSE84133mouse1 & Baron \cite{baron2016single}& Mouse & 6 & 782 & 14878\\
        GSE84133mouse2 & Baron \cite{baron2016single} & Mouse  & 8 & 1036 & 14878\\
        GSE57249 & Biase \cite{biase2014cell} & Human &  3 & 49 & 25737\\
        GSE64016 & Leng \cite{leng2015oscope} & Human &  4 & 460 & 19084\\
        GSE94820 & Villani \cite{villani2017single} & Human & 5 & 1140  & 26593\\\hline
    \end{tabular}
    \label{tab: data}
\end{table}

\subsection{Benchmarking PL regularized NMF}

In order to benchmark persistent Laplacain regularized NMF, we compared our methods to other commonly used NMF methods, namely the GNMF, rGNMF, rNMF and NMF. For a fair comparison,  We omitted  supervised or semi-supervised methods. For $k$-rTNMF, rTNMF, $k$-TNMF, TNMF, GNMF and rGNMF, we set $\alpha = 1$ for all   tests.

\begin{figure}
    \centering
    \includegraphics{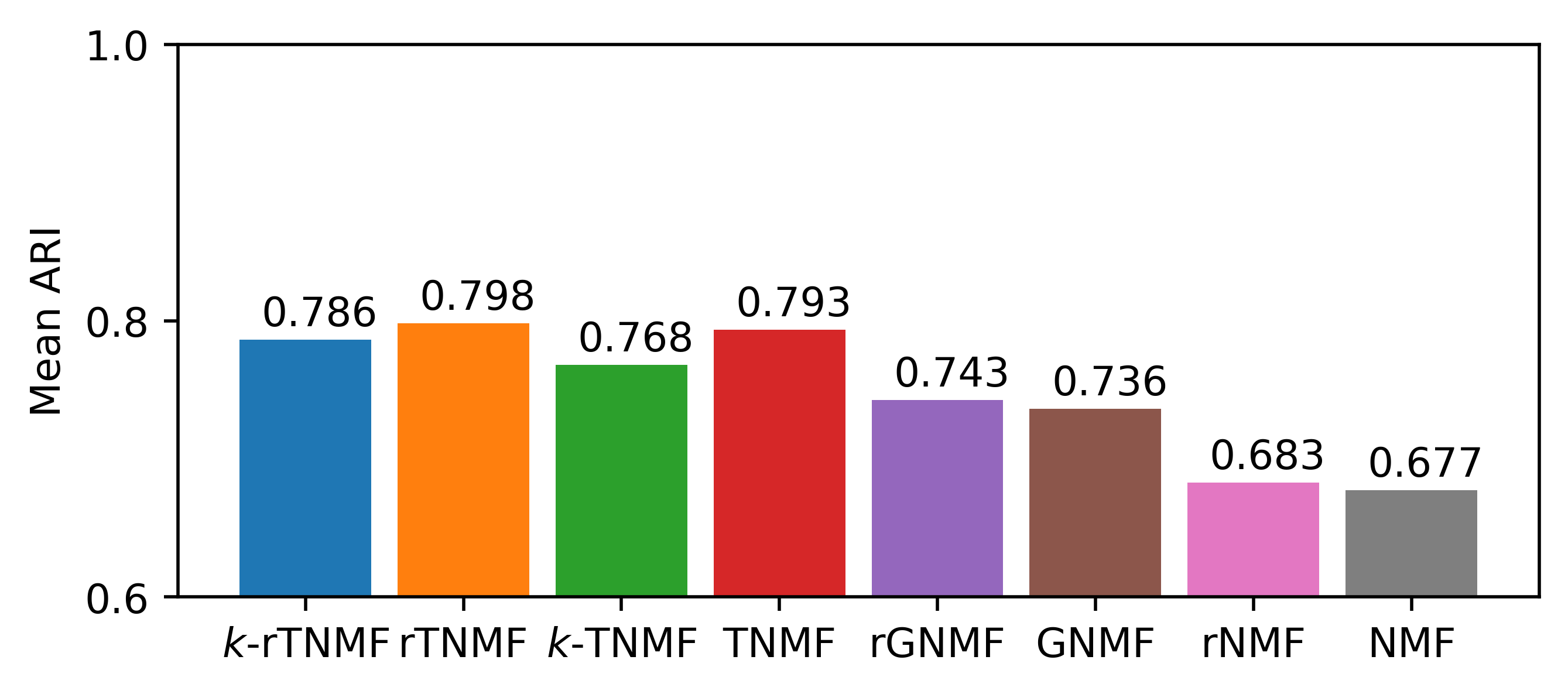}
    \caption{Average ARI of $k$-rTNMF, rTNMF, $k$-TNMF,TNMF, rGNMF, GNMF, rNMF and NMF for the 12 datasets}
    \label{fig: ARI}
\end{figure}

\autoref{tab: ARI} shows the ARI values  of the NMF methods for the 12 data we have tested. The bold number indicate the highest performance. \autoref{fig: ARI} depicts the average ARI value  over the 12 datasets for each method.

\begin{table}[H] 
    \caption{ARI of NMF methods across 12 datasets.}
        \begin{tabular}{|c | c c c c c c c c |} \hline  
        data & $k$-rTNMF  & rTNMF & $k$-TNMF & TNMF & rGNMF & GNMF & rNMF & NMF  \\ \hline
        GSE67835 & \textbf{0.9454} & 0.9236 & 0.9306 & 0.8533 & 0.9391 & 0.9109 & 0.7295 & 0.7314\\
        GSE64016 & \textbf{0.2569} & 0.1544 & 0.2237 & 0.1491 & 0.1456 & 0.1605 & 0.1455 & 0.1466\\
        GSE75748time & 0.6421 & \textbf{0.6581} & 0.5963 & 0.6099 & 0.6104 & 0.5790 & 0.5969 & 0.5996\\
        GSE82187 & \textbf{0.9877} & 0.9815 & 0.9676 & 0.9809 & 0.7558 & 0.7577 & 0.8221 & 0.8208\\
        GSE84133human1 & 0.8310 & \textbf{0.8969} & 0.8301 & 0.8855 & 0.8220 & 0.7907 & 0.7080 & 0.6120\\
        GSE84133human2 & \textbf{0.9469} & 0.9072 & 0.9433 & 0.9255 & 0.9350 & 0.9255 & 0.8930 & 0.8929\\
        GSE84133human3 & 0.8504 & 0.9179 & 0.8625 & \textbf{0.9181} & 0.8447 & 0.8361 & 0.7909 & 0.8089\\
        GSE84133human4 & 0.8712 & \textbf{0.9692} & 0.8712 & 0.9692 & 0.8699 & 0.8681 & 0.8311 & 0.8311\\
        GSE84133mouse1 & \textbf{0.8003} & 0.7894 & 0.8003 & 0.7913 & 0.7945 & 0.7918 & 0.6428 & 0.6348\\
        GSE84133mouse2 & 0.6953 & 0.8689 & 0.7005 & \textbf{0.9331} & 0.6808 & 0.6957 & 0.5436 & 0.5470\\
        GSE57249 & \textbf{1.0000} & 0.9638 & 1.0000 & 0.9483 & 1.0000 & 1.0000 & 0.9483 & 0.9483\\
        GSE94820 & \textbf{0.6101} & 0.5480 & 0.4916 & 0.5574 & 0.5139 & 0.5189 & 0.5440 & 0.5556\\ \hline
        \end{tabular}
    \label{tab: ARI}
\end{table}

Overall, PL regularized rNMF and NMF have the highest ARI value across all the datasets. $k$-rTNMF outperforms other NMF methods by at least 0.09 for GSE64016. All PL regularized NMF methods outperform other NMF methods by at least 0.14 for GSE82187.  For GSE84133 human 3, both rTNMF and TNMF outperform other methods by 0.07. TNMF improves other methods by more than 0.2 for GSE84133 mouse 2. Lastly, $k$-rTNMF has the highest ARI value for GSE94820. Moreover, rTNMF improves rGNMF by 0.05, and TNMF improves GNMF by about 0.06. $k$-TNMF and $k$-rTNMF also improve GNMF and rGNMF by about 0.03.

\begin{figure}[H]
    \centering
    \includegraphics{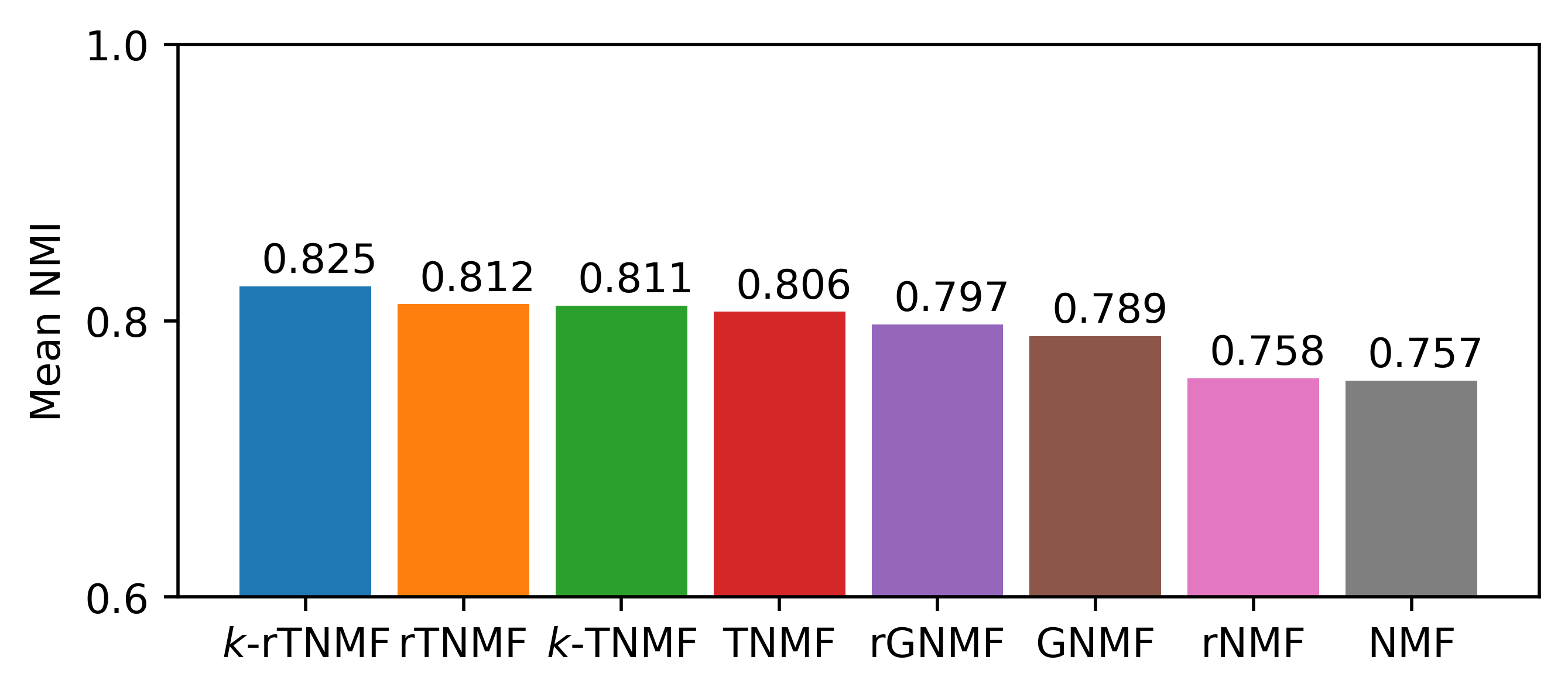}
    \caption{Average NMI values of $k$-rTNMF, rTNMF, $k$-TNMF,TNMF, rGNMF, GNMF, rNMF and NMF for the 12 datasets}
    \label{fig: NMI}
\end{figure}

\autoref{tab: NMI} shows the NMI values of of the NMF methods for the 12 datasets we have tested. The bold  number indicate the highest performance. \autoref{fig: NMI} shows the average NMI value over the 12 datasets.
\begin{table}[H] 
    \caption{NMI of NMF methods across 12 datasets.}
        \begin{tabular}{|c | c c c c c c c c|} \hline  
        data & $k$-rTNMF  & rTNMF & $k$-TNMF & TNMF & rGNMF & GNMF & rNMF & NMF  \\ \hline
        GSE67835 & \textbf{0.9235} & 0.8999 & 0.9107 & 0.8607 & 0.9104 & 0.8858 & 0.7975 & 0.8017\\
        GSE64016 & 0.\textbf{3057} & 0.2059 & 0.3136 & 0.1869 & 0.2593 & 0.2562 & 0.1896 & 0.1849\\
        GSE75748time & 0.7522 & \textbf{0.7750} & 0.7159 & 0.7343 & 0.7235 & 0.6971 & 0.7227 & 0.7244\\
        GSE82187 & \textbf{0.9759} & 0.9691 & 0.9298 & 0.9668 & 0.8802 & 0.8754 & 0.9124 & 0.9117\\
        GSE84133human1 & \textbf{0.8802} & 0.8716 & 0.8785 & 0.8780 & 0.8713 & 0.8310 & 0.8226 & 0.7949\\
        GSE84133human2 & \textbf{0.9363} & 0.8937 & 0.9313 & 0.9070 & 0.9237 & 0.9145 & 0.8835 & 0.8829\\
        GSE84133human3 & 0.8500 & \textbf{0.8718} & 0.8577 & 0.8677 & 0.8439 & 0.8357 & 0.8215 & 0.8260\\
        GSE84133human4 & 0.8795 & \textbf{0.9542} & 0.8795 & 0.9542 & 0.8775 & 0.8753 & 0.8694 & 0.8694\\
        GSE84133mouse1 & \textbf{0.8664} & 0.8498 & 0.8664 & 0.8495 & 0.8596 & 0.8565 & 0.7634 & 0.7593\\
        GSE84133mouse2 & 0.8218 & \textbf{0.8355} & 0.8299 & 0.8713 & 0.8005 & 0.8129 & 0.7258 & 0.7272\\
        GSE57249 & \textbf{1.0000} & 0.9505 & 1.0000 & 0.9293 & 1.0000 & 1.0000 & 0.9293 & 0.9293\\
        GSE94820 &\textbf{ 0.7085} & 0.6657 & 0.6157 & 0.6716 & 0.6195 & 0.6258 & 0.6624 & 0.6693\\ \hline
    \end{tabular}
    \label{tab: NMI}
\end{table}

Interestingly, $k$-rTNMF and $k$-TNMF on average have   higher NMI values than rTNMF and TNMF, respectively. However, all PL regularized methods outperform rGNMF, GNMF, rNMF and NMF. Overall, PL regularized methods outperform other methods. Most noticeably, $k$-rTNMF, rTNMF and TNMF outperform standard NMF methods by 0.06 for GSE82187. Both rTNMF and TNMf outperform rGNMF and GNMF by 0.08 for GSE84133 human 4.

\begin{figure}[H]
    \centering
    \includegraphics{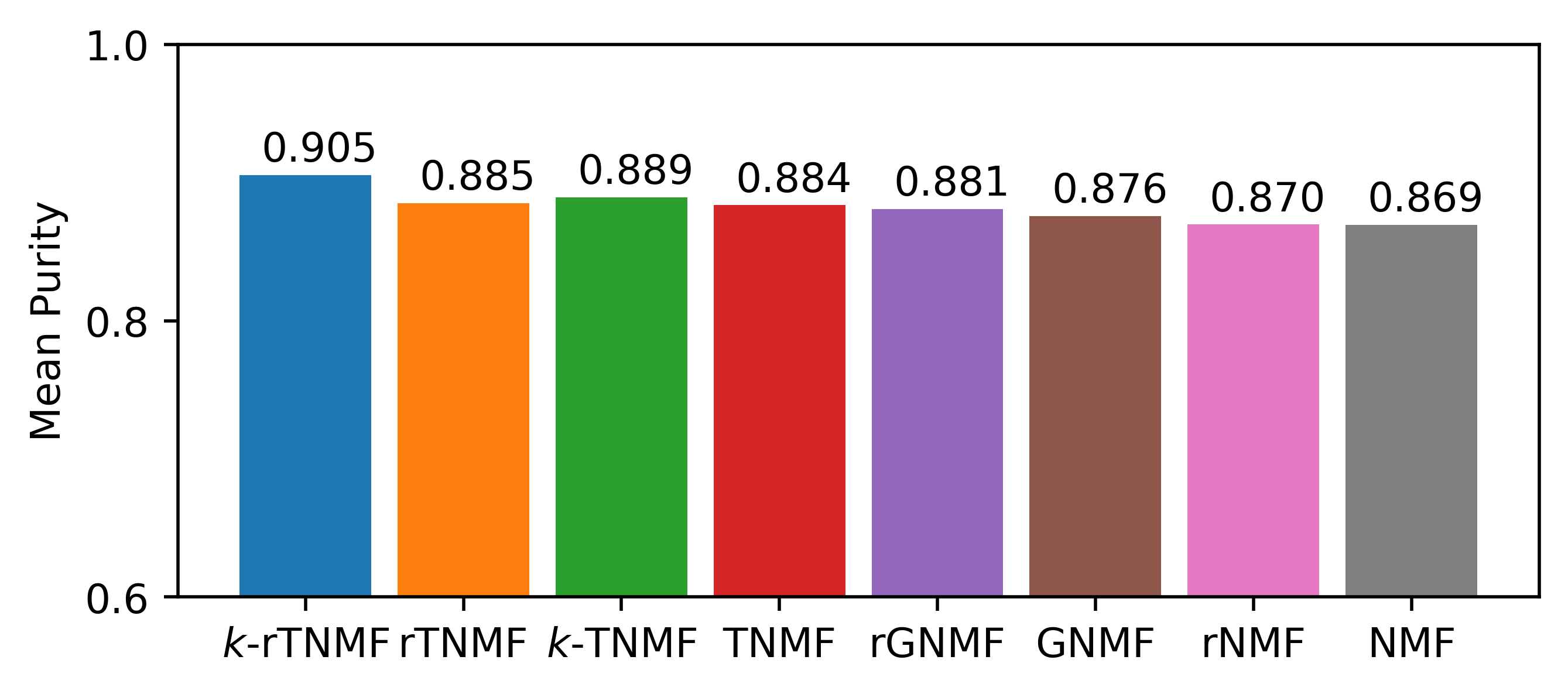}
    \caption{Average purity values of $k$-rTNMF, rTNMF, $k$-TNMF,TNMF, rGNMF, GNMF, rNMF and NMF for the 12 datasets}
    \label{fig: Purity}
\end{figure}

\autoref{tab: Purity} shows the purity values of  the NMF methods for the 12 datasets we have tested. The bold  number indicate the highest performance. \autoref{fig: Purity} shows the average purity over the 12 datasets.

\begin{table}[H] 
    \caption{Purity of NMF methods across 12 datasets.}
        \begin{tabular}{|c | c c c c c c c c|} \hline  
        data & $k$-rTNMF  & rTNMF & $k$-TNMF & TNMF & rGNMF & GNMF & rNMF & NMF  \\ \hline
        GSE67835 & \textbf{0.9643} & 0.9267 & 0.9595 & 0.9024 & 0.9595 & 0.9476 & 0.8726 & 0.8719\\
        GSE64016 & \textbf{0.6048} & 0.4913 & 0.5846 & 0.5013 & 0.5339 & 0.5398 & 0.5080 & 0.5050\\
        GSE75748time & \textbf{0.7736} & 0.7512 & 0.7533 & 0.7454 & 0.7553 & 0.7387 & 0.7467 & 0.7455\\
        GSE82187 & \textbf{0.9927} & 0.9895 & 0.9620 & 0.9888 & 0.9620 & 0.9594 & 0.9693 & 0.9692\\
        GSE84133human1 & \textbf{0.9543} & 0.9357 & 0.9536 & 0.9382 & 0.9490 & 0.9187 & 0.9189 & 0.9099\\
        GSE84133human2 & \textbf{0.9818} & 0.9614 & 0.9806 & 0.9661 & 0.9777 & 0.9736 & 0.9602 & 0.9600\\
        GSE84133human3 & 0.9472 & \textbf{0.9485} & 0.9531 & 0.9460 & 0.9452 & 0.9420 & 0.9464 & 0.9466\\
        GSE84133human4 & 0.9427 & \textbf{0.9882} & 0.9427 & 0.9882 & 0.9427 & 0.9420 & 0.9412 & 0.9412\\
        GSE84133mouse1 & \textbf{0.9565} & 0.9540 & 0.9565 & 0.9540 & 0.9552 & 0.9540 & 0.9309 & 0.9299\\
        GSE84133mouse2 & 0.9585 & 0.9410 & \textbf{0.9604} & 0.9373 & 0.9466 & 0.9507 & 0.9185 & 0.9199\\
        GSE57249 & \textbf{1.0000} & 0.9857 & 1.0000 & 0.9796 & 1.0000 & 1.0000 & 0.9796 & 0.9796\\
        GSE94820 & \textbf{0.7893} & 0.7462 & 0.6658 & 0.7550 & 0.6421 & 0.6421 & 0.7429 & 0.7531\\ \hline
    \end{tabular}
    \label{tab: Purity}
\end{table}

In general, PL-regularized methods achieve higher purity values compared to other NMF methods. Purity measures the maximum intersection between true and predicted classes, which is why we do not observe a significant difference, as seen in ARI and NMI. Furthermore, since purity does not account for the size of a class, and given the imbalanced class sizes in scNRA-seq data, it is not surprising that the purity values are similar.

\begin{figure}[H]
    \centering
    \includegraphics{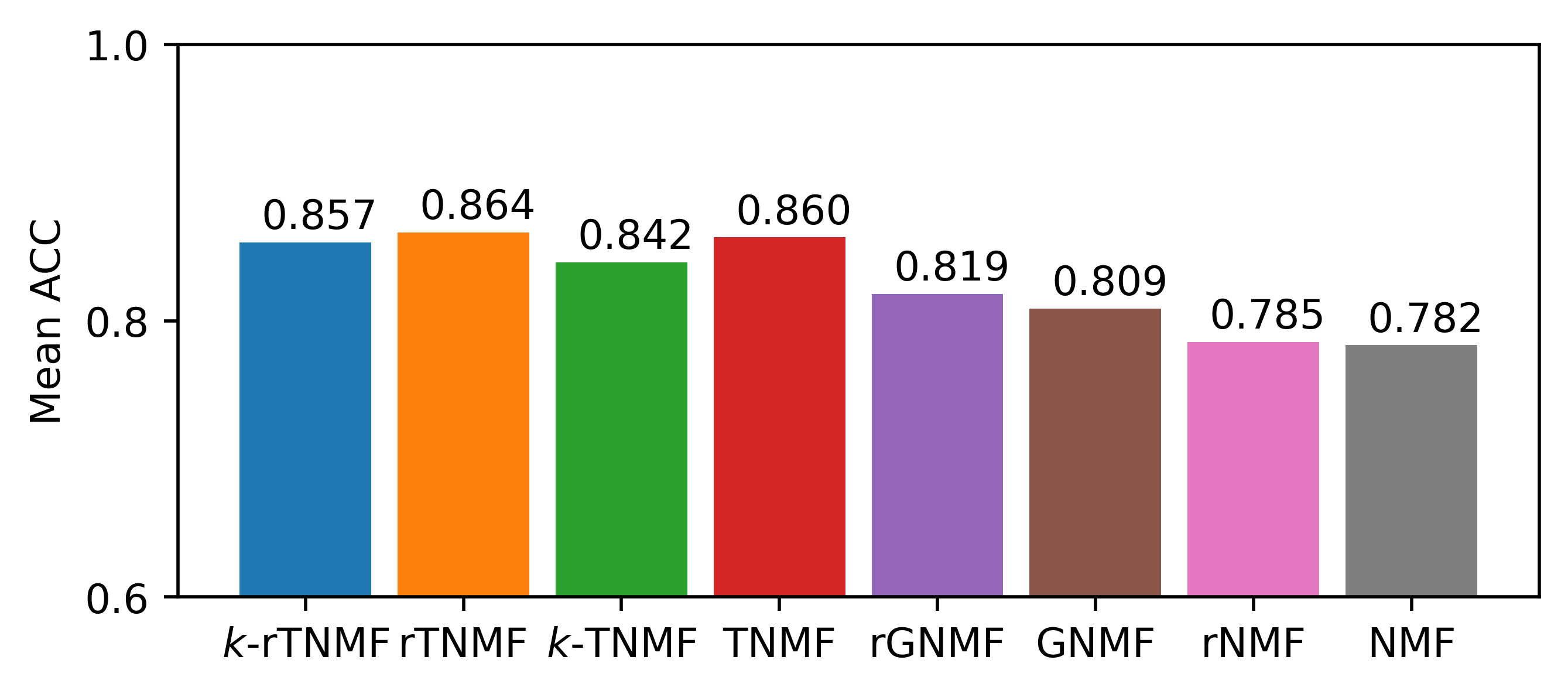}
    \caption{Average ACC of $k$-rTNMF, rTNMF, $k$-TNMF,TNMF, rGNMF, GNMF, rNMF and NMF for the 12 datasets}
    \label{fig: ACC}
\end{figure}

\autoref{tab: ACC} shows the ACC of   the NMF methods for the 12 datasets we have tested. The bold number indicate the highest performance. \autoref{fig: ACC} shows the average ACC over the 12 datasets.
\begin{table}[H] 
    \caption{ACC of NMF methods across 12 datasets.}
        \begin{tabular}{|c | c c c c c c c c|} \hline  
        data & $k$-rTNMF  & rTNMF & $k$-TNMF & TNMF & rGNMF & GNMF & rNMF & NMF  \\ \hline
        GSE67835 & \textbf{0.9643} & 0.9243 & 0.9595 & 0.9000 & 0.9595 & 0.9383 & 0.8357 & 0.8364\\
        GSE64016 & \textbf{0.5700} & 0.4870 & 0.5502 & 0.4746 & 0.4891 & 0.4537 & 0.4691 & 0.4759\\
        GSE75748time & \textbf{0.7565} & 0.7438 & 0.7414 & 0.6917 & 0.7355 & 0.7241 & 0.6873 & 0.6875\\
        GSE82187 & \textbf{0.9927} & 0.9895 & 0.9599 & 0.9888 & 0.8512 & 0.8514 & 0.8896 & 0.8889\\
        GSE84133human1 & 0.8973 & \textbf{0.9194} & 0.8974 & 0.9088 & 0.8889 & 0.8364 & 0.7988 & 0.7370\\
        GSE84133human2 & \textbf{0.9260} & 0.9069 & 0.9242 & 0.9447 & 0.9224 & 0.9177 & 0.8998 & 0.8994\\
        GSE84133human3 & 0.8539 & \textbf{0.9456} & 0.8597 & 0.9419 & 0.8498 & 0.8228 & 0.8032 & 0.8178\\
        GSE84133human4 & 0.8831 & \textbf{0.9882} & 0.8831 & 0.9882 & 0.8824 & 0.8816 & 0.8847 & 0.8847\\
        GSE84133mouse1 & \textbf{0.8581} & 0.8542 & 0.8581 & 0.8542 & 0.8555 & 0.8542 & 0.7361 & 0.7311\\
        GSE84133mouse2 & 0.8232 & \textbf{0.9101} & 0.8263 & 0.9305 & 0.7903 & 0.8155 & 0.7239 & 0.7294\\
        GSE57249 & \textbf{1.0000} & 0.9857 & 1.0000 & 0.9796 & 1.0000 & 1.0000 & 0.9796 & 0.9796\\
        GSE94820 & \textbf{0.7533} & 0.7119 & 0.6482 & 0.7201 & 0.6088 & 0.6107 & 0.7091 & 0.7189\\ \hline
        \end{tabular}
    \label{tab: ACC}
\end{table}

Once again, we see that PL regularized methods have higher ACC than other NMF methods. RTNMF and TNMF improves rGNMF and GNMF by 0.05, and $k$-rTNMF and $k$-TNMF improves rGNMF and GNMF by 0.04. We see an improvement in ACC for both $k$-rTNMF and $k$-TNMF for GSE64016. All 4 PL regularized methods improve ACC of GSE82187 by 0.1. RTNMF and TNMF improve GSE84133 mouse 2 by at least 0.1 as well.

%From the benchmark, we observe the following:
%\begin{enumerate}[label=(\alph*)]
%    \item RNMF has slightly better performance than the standard NMF, which is consistent with other literatures. In particular,rNMF performs notably better in all the metric for GSE84133 human 1 data.
%    \item RGNMF and GNMF have comparable result. Most notably, rGNMF performs better on GSE84133 human 1 data, whereas GNMF perform better on GSE84133 mouse 2 data.
%    \item In general PL regularized NMF outperform other methods.
%    \item $k$-rTNMF and $k$-TNMF has a better overall performance than rTNMF and TNMF, respectively. Specifically, for GSE67835, TNMF have worse performance than GNMF and rGNMF, but the $k$-TNMF outperforms both GNMF and rGNMF. This may be caused by inconsistency in the number of edges in the cutoff based PL. We have a much better control in the number of edges for $k$-NN induced PL.
%    \item There is a considerable improvement in performance for PL regularized NMF for GSE75748 time, GSE82187 and GSE64016 than the GNMf and rGNMF. This exemplifies the drawback of single scale method, ie the graph Laplacian. Since PL allow for flexibility in weighing interactions at different scales, it allows for a better performance.
%    \item There is a considerable improvement in performance in all metric for $k$-rTNMF for GSE94820.
    
%\end{enumerate}

\subsection{Overall performance}
\begin{figure}
    \centering
    \includegraphics{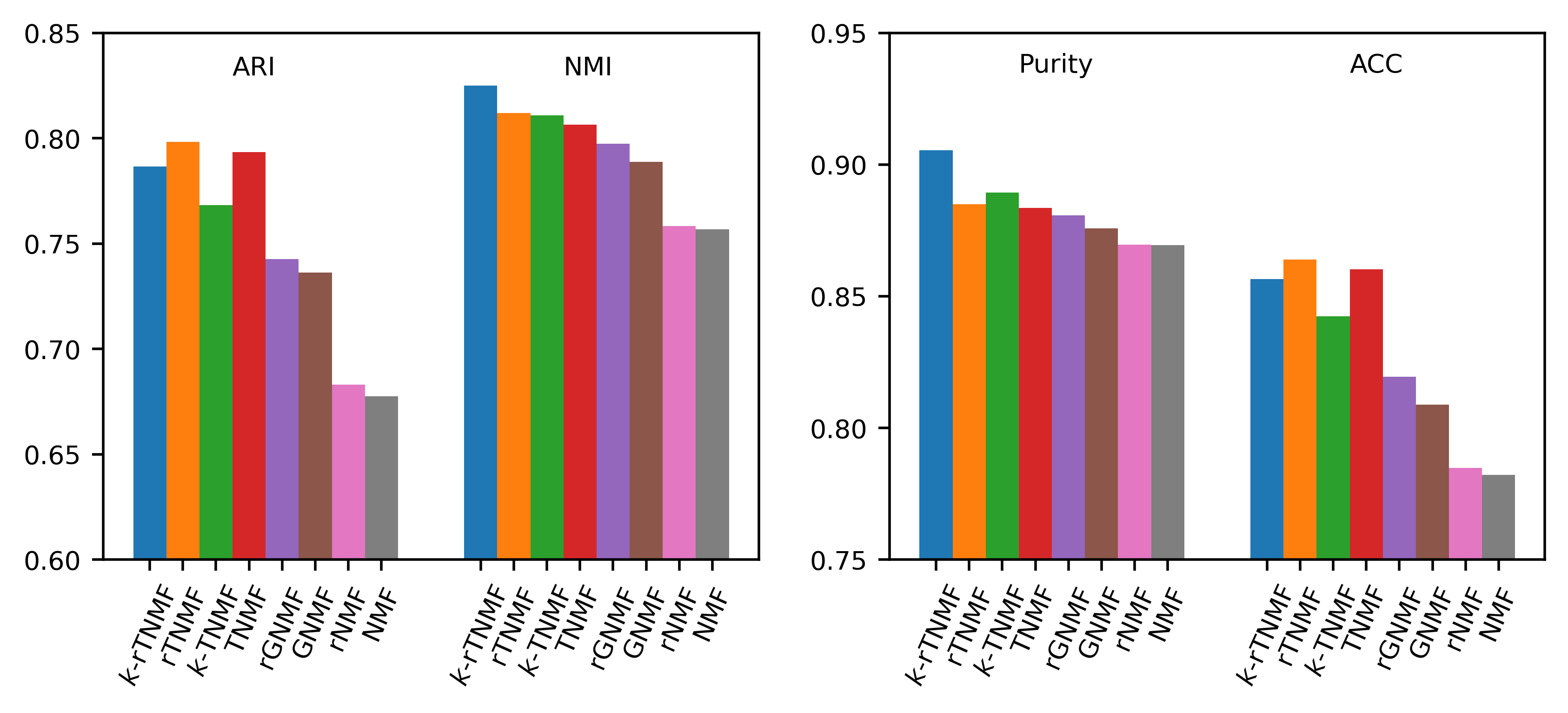}
    \caption{Average ARI, NMI, purity and ACC of $k$-rTNMF, rTNMF, $k$-TNMF, TNMF, rGNMF, GNMF, rNMF, NMF across 10 datasets}
    \label{fig: overall}
\end{figure}

\autoref{fig: overall} shows the average ARI, NMI, purity and ACC of $k$-rTNMF, rTNMF, $k$-TNMF, TNMF, rGNMF, GNMF, rNMF, NMF across 10 datasets. All PL regularized NMF methods  outperform  the traditional rGNMF, GNMF, rNMF and NMF. Both rTNMF and TNMF have higher average ARI and purity than the $k$-NN based PL counterparts. However, $k$-rTNMF and $k$-TNMF have higher average NMI than rTNMF and TNMF, respectively. $k$-rTNMF has a significantly higher purity than other methods.

\section{Discussion}

\subsection{Visualization of meta-genes based UMAP and t-SNE}
Both UMAP and t-SNE are well-known for their effectiveness in visualization. However, these methods may not perform as competitively in clustering or classification tasks. Therefore, it is beneficial to employ NMF-based methods to enhance the visualization capabilities of UMAP and t-SNE.

In this process, we generate meta-genes and subsequently utilize UMAP or t-SNE to further reduce the data to 2 dimensions for visualization. For a dataset with $M$ cells, the number of meta-genes will be the integer value of $\sqrt{M}$. To compare the standard UMAP and t-SNE plots with the top-NMF-assisted and top-rNMF-assisted UMAP and t-SNE visualizations, we used the default settings of the Python implementation of UMAP and the Scikit-learn implementation of t-SNE. For unassisted UMAP and t-SNE, we first removed low-abundance genes and performed log-transformation before applying UMAP and t-SNE.

\begin{figure}[H]
    \centering
    \includegraphics[width = \textwidth]{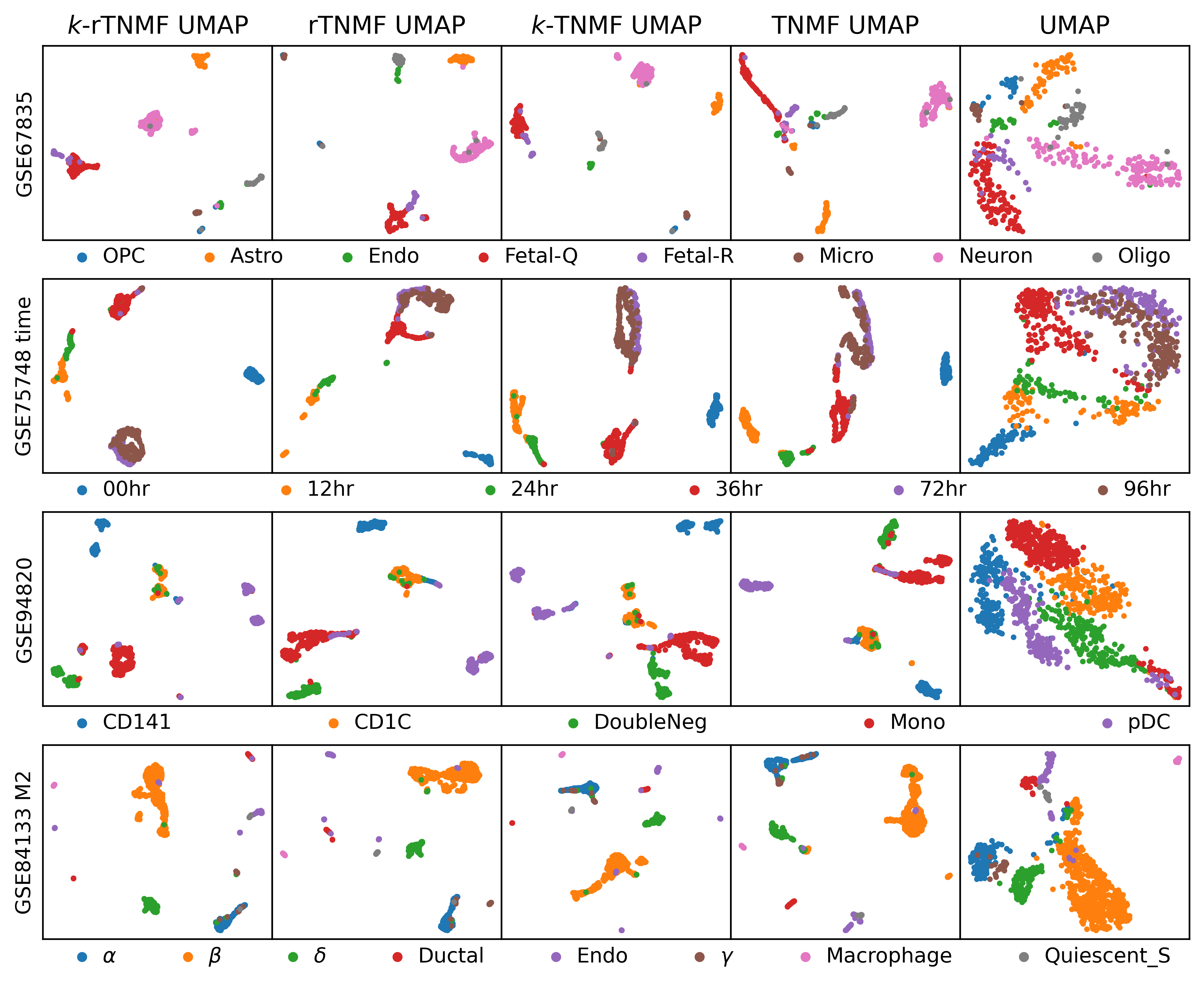}
    \caption{Visualization of top-NMF and top-rNMF meta-genes through UMAP. Each row corresponds to GSE67835, GSE75748 time, GSE94820 and GSE84133 mouse 2 data. The columns from left to right are the $k$-rTNMF assisted UMAP, rTNMF assisted UMAP, $k$-TNMF assisted UMAP, TNMF assisted UMAP and UMAP visualization. Samples were colored according to their true cell types}
    \label{fig: top umap}
\end{figure}

\autoref{fig: top umap} shows the visualization of PL regularized NMF methods  through UMAP. Each row corresponds to GSE67835, GSE75748 time, GSE94820 and GSE84133 mouse 2 data. The columns from left to right are the $k$-rTNMF assisted UMAP, rTNMF assisted UMAP, $k$-TNMF assisted UMAP, TNMF assisted UMAP and UMAP visualization. Samples were colored according to their true cell types.

\begin{figure}[H]
    \centering
    \includegraphics[width = \textwidth]{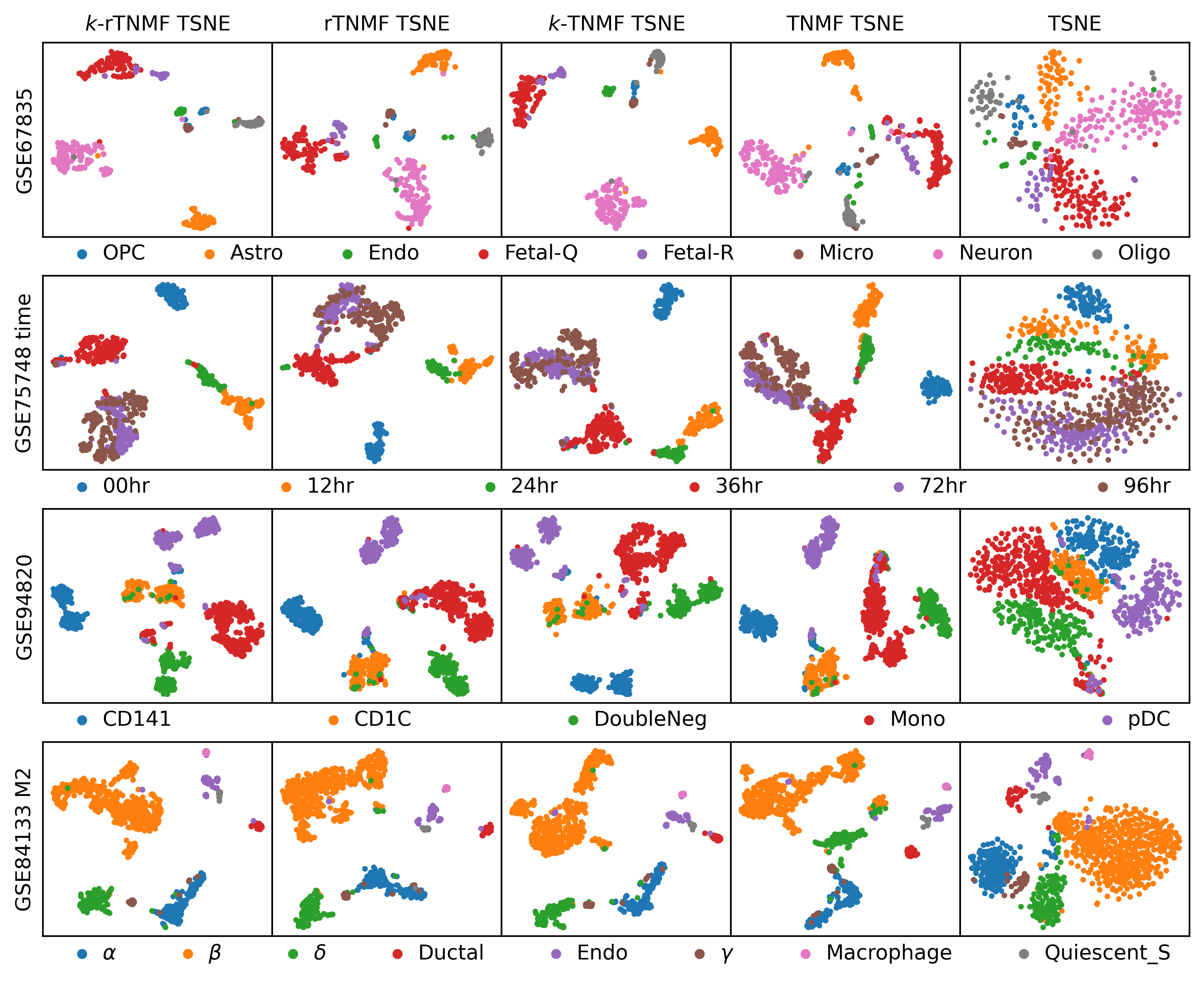}
    \caption{Visualization of top-NMF and top-rNMF meta-genes through t-SNE. Each row corresponds to GSE67835, GSE75748 time, GSE94820 and GSE84133 mouse 2 data. The columns from left to right are the $k$-rTNMF assisted t-SNE, rTNMF assisted t-SNE, $k$-TNMF asssited t-SNE, TNMF assisted t-SNE and t-SNE visualization. Samples were colored according to their true cell types}
    \label{fig: top tsne}
\end{figure}

\autoref{fig: top tsne} shows the visualization of PL regularized NMF through t-SNE. Each row corresponds to GSE67835, GSE75748 time, GSE94820 and GSE84133 mouse 2 data. The columns from left to right are the $k$-rTNMF assisted t-SNE, rTNMF assisted t-SNE, $k$-TNMF assisted t-SNE, TNMF assisted t-SNE and t-SNE visualization. Samples were colored according to their true cell types.

We see a considerable improvement in both top-NMF assisted and top-rNMF assisted UMAP and t-SNE visualization. 

\paragraph{GSE67835}
In the assisted UMAP and t-SNE visualizations of GSE67835, we observe a more distinct cluster, which includes a supercluster of fetal quiescent (Fetal-Q) and fetal replicating (Fetal-R) cells. Darmanis et al. \cite{darmanis2015survey} conducted a study that involved obtaining differential gene expression data for human adult brain cells and sequencing fetal brain cells for comparison. It is not surprising that the undeveloped Fetal-Q and Fetal-R cells do not exhibit significant differences and cluster together.

\paragraph{GSE75748 time}
In GSE75748 time data, Chu et al. \cite{chu2016single} sequenced human embryonic stem cells at times 0hr, 12hr, 24hr, 36hr, 72hr, and 96hr under hypoxic conditions to observe differentiation. In unassisted UMAP and t-SNE, although some clustering is visible, there is no clear separation between the clusters. Additionally, two subclusters of 12hr cells are observed.

Notably, in the PL-regularized assisted UMAP and t-SNE visualizations, there is a distinct supercluster comprising the 72hr and 96hr cells, while cells from different time points form their own separate clusters. This finding aligns with Chu's observation that there was no significant difference between the 72hr and 96hr cells, suggesting that differentiation may have already occurred by the 72hr mark.

\paragraph{GSE94820}
Notice that in both t-SNE and UMAP, although there is a boundary, the cells do not form distinct clusters. This lack of distinct clustering can pose challenges in many clustering and classification methods. On the other hand, all PL-regularized NMF methods result in distinct clusters.

Among the PL-regularized NMF approaches, cutoff-based PL, rTNMF, and TNMF form a single CD1C$^+$ (CD1C1) cluster, whereas the $k$-NN induced PL, $k$-rTNMF, and $k$-TNMF exhibit two subclusters. Villani et al. \cite{villani2017single} previously noted the similarity in the expression profile of CD1C1$^-$CD141$^-$ (DoubleNeg) cells and monocytes. PL-regularized NMF successfully differentiates between these two types.

\paragraph{GSE84133 mouse 2}
PL-regularized NMF yields significantly more distinct clusters compared to unassisted UMAP and t-SNE. Notably, the beta and gamma cells form distinct clusters in PL-regularized NMF. Additionally, when PL-regularized NMF is applied to assist UMAP, potential outliers within the beta cell population become visible. Baron et al. \cite{baron2016single} previously highlighted heterogeneity within the beta cell population, and we observe potential outliers in all visualizations.

\subsection{RS analysis}
Although UMAP and t-SNE are excellent tools for visualizing clusters, they may struggle to capture heterogeneity within clusters. Moreover, these methods can be less effective when dealing with a large number of classes. Therefore, it is essential to explore alternative visualization techniques.

\begin{figure}
    \centering
    \includegraphics{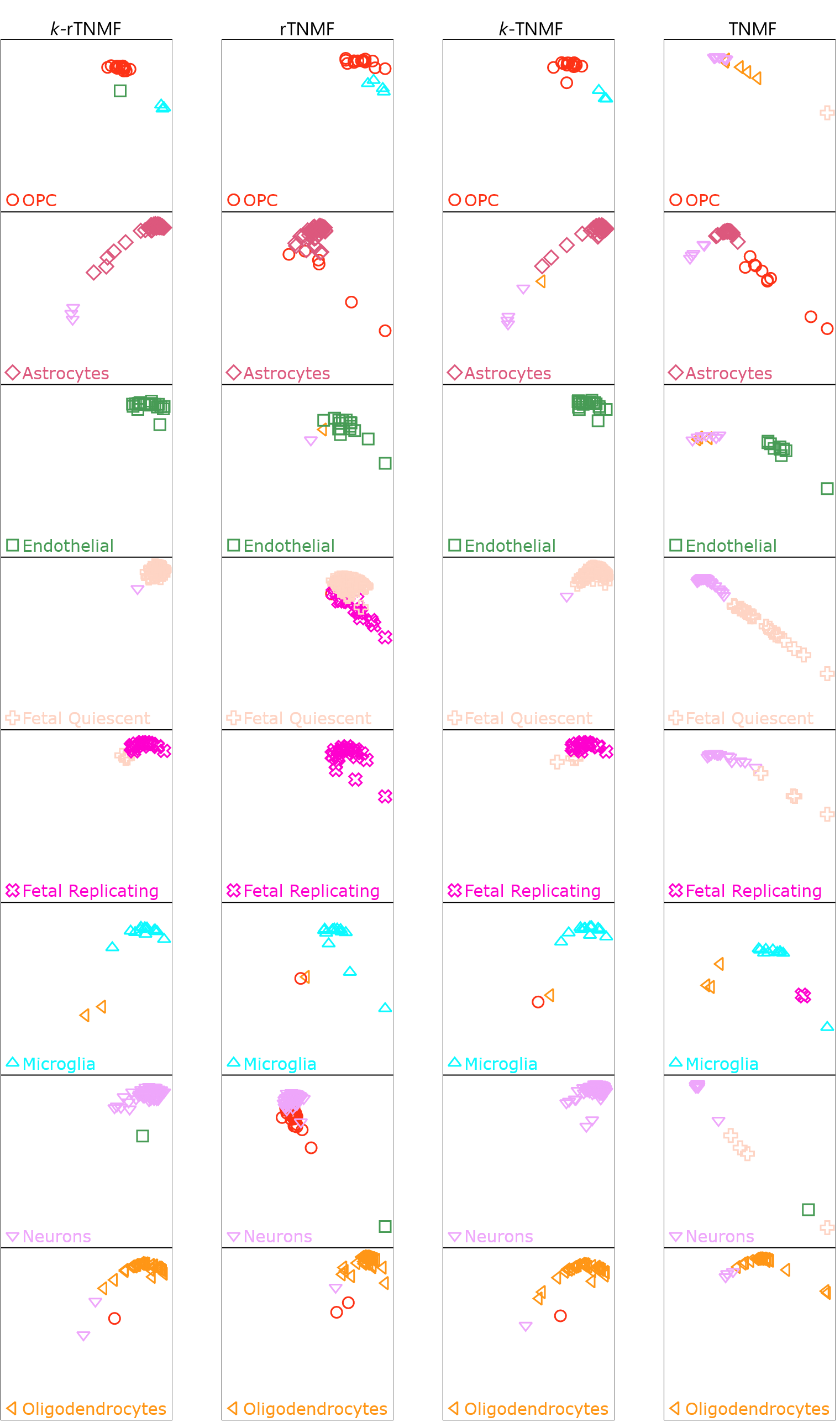}
    \caption{RS plots of GSE67835 data. The columns from left to right correspond to $k$-rTNMF, rTNMF, $k$-TNMF, and TNMF. Each row corresponds to a cell type. For each section, the x-axis and y-axis correspond to the S-score and R-score, respectively. K-means was used to obtain a cluster label, and the Hungarian algorithm was used to map the cluster labels to the true labels. Each sample was colored according to their true labels.}
    \label{fig: RS plot}
\end{figure}
In our approach, we visualize each cluster using RS plots \cite{hozumi2022ccp}. RS plots depict the relationship between the residue score (R score) and similarity score (S score) and have proven useful in various applications for visualizing data with multiple class types \cite{hozumi2023preprocessing, feng2023virtual, zhu2023tidal, shen2023svsbi, cottrell2023plpca}. 

Let $\{(\mathbf{x}_m, y_m) | \mathbf{x}_m \in  \mathbb{R}^N, y_m \in \mathbb{Z}_L, 1 \le m \le M\}$ be the data, where $\mathbf{x}_m$ is the $m$th sample, $y_m$ is the cell type or cluster label. $L$ is the number of class. That is, $C_l = \{\mathbf{x}_m \in \mathcal{X} | y_m = l\}$ and $\uplus_{0}^{L-1} \mathcal{C}_l = \mathcal{X}$.
	
The  residue (R) score is defined as the inter-class sum of distance. For a given data $\textbf{x}_m$ with assignment $y_m = l$, the R-score is defined as 
\begin{align*}
    & R_m = R(\mathbf{x}_m) = \frac{1}{R_{\max}}\sum_{\mathbf{x}_j \notin \mathcal{C}_l} \|\mathbf{x}_m - \mathbf{x}_j\|,
\end{align*}
where $\displaystyle R_{\max} = \max_{\mathbf{x}_m, \mathbf{x}_m\in \mathcal{X}} R_m$. The similarity (S) score is defined as the intra-class average of distance, defined as 
\begin{align*}
    S_m = S(\mathbf{x}_m) = \frac{1}{|\mathcal{C}_l|}\sum_{\mathbf{x}_j \in \mathcal{C}_l} \left(1 - \frac{\|\mathbf{x}_m - \mathbf{x}_j\|}{d_{\max}}\right),
\end{align*}
where $\displaystyle d_{\max} = \max_{\mathbf{x}_i, \mathbf{x}_j \in \mathcal{X}} \|\mathbf{x}_i - \mathbf{x}_j\|$ and $|\mathcal{C}_l|$ is the number of data in class $\mathcal{C}_l$. Both $R_m$ and $S_m$ are bounded by 0 and 1,  and the larger the better for a given dataset.

The class residue index (CRI) and the class similarity index (CSI) can then be defined as the average of the R-score and S-score of each of the classes. That is $ \text{CRI}_l = \frac{1}{|\mathcal{C}_l|}\sum_{m} R_m$ and $ \text{CSI}_l = \frac{1}{|\mathcal{C}_l|}\sum_{ m  } S_m$.
Then, the residue index (RI) and the similarity index (SI) can be defined $ \text{RI} = \frac{1}{L}\text{CRI}_l$ and $ \text{SI} = \frac{1}{L}\text{CSI}_l $, respectively.

Using the RI and SI, the residue similarity disparity can be computed by taking $\text{RSD} = \text{RI} - \text{SI}$, and the residue-similarity index (RSI) can be computed as $\displaystyle \text{RSI} = 1 - |\text{RI} - \text{SI}|$.

\autoref{fig: RS plot} shows the RS plots of PL-regularized NMF methods for GSE67835 data. The columns from left to right correspond to $k$-rTNMF, rTNMF, $k$-TNMF, and TNMF, while the rows correspond to the cell types. The x-axis and y-axis represent the S-score and R-score for each sample, respectively. The samples are colored according to their predicted cell types. Predictions were obtained using k-means clustering, and the Hungarian algorithm was employed to find the optimal mapping from the cluster labels to the true cell types.

We can see that TNMF fails to identify OP cells, whereas $k$-rTNMF, rTNMF, and $k$-TNMF are able to identify OPC cells. Notably, the S-score is quite low, indicating that the OPC did not form a cluster for TNMF. For fetal quiescent and replicating cells, $k$-rTNMF correctly identifies these two types, and the few misclassified samples are located on the boundaries. RTNMF is able to correctly identify fetal replicating cells but could not distinguish fetal quiescent cells from fetal replicating cells. The S-score is low for neurons in both rTNMF and TNMF, which shows a direct correlation with the number of misclassified cells.

\section{Conclusion}

Persistent Laplacian-regularized NMF is a dimensionality reduction technique that incorporates multiscale topological interactions between the cells. Traditional graph Laplacian-based regularization only represents a single scale and cannot capture the multiscale features of the data. We have also shown that the $k$-NN induced persistent Laplacian outperforms other NMF methods and is comparable to the cutoff-based persistent Laplacian-regularized NMF methods. However, PL methods do come with their downside. In particular, the weights for each filtration must be determined prior to the reduction. If there are $T$ filtrations, then the hyperparameter space is $(T+1)^T$. However, $k$-NN induced PL reduces the number of parameters to $2^T$. In addition, we have shown that we can achieve a significant improvement even if we limit the hyperparameter space to $2^T$. We would like to further explore possible parameter-free versions of topological NMF. Additionally, NMF methods are not globally convex, but we have shown that with NNDSVDA initialization, our methods perform the best. One possible extension to the proposed methods is to incorporate higher-order persistent Laplacians in the regularization framework, which will reveal higher-order interactions. In addition, we would like to expand the ideas to tensor decomposition, such as Canonical Polyadic Decomposition (CPD) and Tucker decomposition, multimodal omics data, and spatial transcriptomics data.

\section{Data availability and code}
The data and model used to produce these results can be obtained at \href{https://github.com/hozumiyu/TopologicalNMF-scRNAseq}{https://github.com/hozumiyu/TopologicalNMF-scRNAseq}.

\section{Acknowledgment}
This work was supported in part by NIH grants R01GM126189, R01AI164266, and R35GM148196, National Science Foundation grants DMS2052983, DMS-1761320, and IIS-1900473, NASA  grant 80NSSC21M0023,   Michigan State University Research Foundation, and  Bristol-Myers Squibb  65109.

\clearpage 

\bibliographystyle{unsrt}
%% \bibliographystyle{custom}

% \end{multicols}

\end{document}